\documentclass{article} 
\usepackage{spconf,amsmath,graphicx}
\usepackage{caption}
\usepackage{amsfonts}
\usepackage{subcaption}
\usepackage{fancyhdr}

\title{FASTER UNSUPERVISED SEMANTIC INPAINTING: A GAN BASED APPROACH}
\name{Avisek Lahiri$^{1 \dagger \ddagger}$ \thanks{$^ \dagger$Funded by Google PhD Fellowship \& Qualcomm Fellowship} \thanks{$^\ddagger$Denotes equal contribution}\qquad Arnav Kumar Jain$^{2* \ddagger}$ \thanks{$^*$Work done while at IIT Kharagpur} \qquad Divyasri Nadendla $^{3*}$ \qquad Prabir Kumar Biswas$^1$}

\address{$^1$ Dept. of E\&ECE, Indian Institute of Technology, Kharagpur, India\\
$^{2}$ Microsoft, India \qquad $^{3}$ Qualcomm, India\\
avisek@ece.iitkgp.ac.in, arnavkj95@gmail.com, divyasri.vss@gmail.com, pkb@ece.iitkgp.ac.in}
\begin{document}
%\ninept
%
\maketitle
\begin{abstract} 
In this paper, we propose to improve the inference speed and visual quality of contemporary baseline of Generative Adversarial Networks (GAN) based unsupervised semantic inpainting. This is made possible with better initialization of the core iterative optimization involved in the framework. To our best knowledge, this is also the first attempt of GAN based video inpainting with consideration to temporal cues. On single image inpainting, we achieve about 4.5-5$\times$ speedup and 80$\times$ on videos compared to baseline. Simultaneously, our method has better spatial and temporal reconstruction qualities as found on three image and one video dataset.
\end{abstract}
\begin{keywords}
Generative Adversarial Networks, Semantic Inpainting, Temporal Consistency, Video Inpainting
\end{keywords}
\section{Introduction}
\label{sec:intro}
Semantic inpainting refers to filling up of missing pixels in a given image by leveraging neighborhood information. Traditional methods \cite{barnes2009patchmatch,hays2007scene} were mainly successful when deployed on background scenes and images with repeated textures. However, they fail to learn complex semantic representations and thereby manifest unpleasing reconstructions on complex non-repetitive textured objects. With the advent of Generative Adversarial Networks (GAN), there has been a recent surge of interest \cite{yeh2017semantic, iizuka2017globally, gip, gfc} to solve inpainting with deep generative models. There are mainly two schools of approach:\\
\textbf{Fully unsupervised:} This approach, first proposed by Yeh \textit{et al.}   \cite{yeh2017semantic} aligns with the concept of the pioneering paper of GAN \cite{goodfellow2014generative}. In \cite{yeh2017semantic}, the objective is learn a GAN model to generate realistic images conditioned on noise priors only and inpaiting is done by iteratively matching a masked/damaged image to its `best matching' noise prior. This method does not require any paired training set (masked, unmasked) and hence we term it as `unsupervised'.\\
\textbf{Hybrid:} These methods \cite{gip, gfc, context_encoders, iizuka2017globally} in general rely on initial training with usual reconstruction loss on a paired (masked, unmasked) dataset. Since $L_2$ loss based reconstructions manifest lack of high frequency components, the next 
 step is to push the solutions nearer to original data manifold with an additional adversarial loss. It is to be noted, that without the initial supervised training phase, these methods fail to work and thus we term this framework as `hybrid' approach.\\
\textbf{Motivation:} For \cite{yeh2017semantic}, being fully unsupervised comes at a cost of significant inference time due to an iterative search for a matching noise prior. Hybrid methods perform test time inference in one single forward pass and thus research has been dedicated mainly towards this genre. However, going against the trend, we advocate the former method because true potential of GAN is appreciated when there is no source of supervision. The motivation in this paper is to primarily reduce the inference run time of \cite{yeh2017semantic}, yet achieve better/similar reconstruction performance compared to \cite{yeh2017semantic}. Towards this, the paper presents the following contributions:

    \textbf{1.} A better initializing method for the  iterative optimization of \cite{yeh2017semantic} to speedup inference run time on single image inpainting by 4.5-5$\times$.
    
     \textbf{2.} First demonstration of totally unsupervised GAN based inpainting on videos (in context of error concealment) with speedup upto 80$\times$ by leveraging temporal redundancy 
    
     \textbf{3.} A group consistency loss for a more temporally consistent sequence reconstruction and thereby leading to more pleasing spatio-temporal experience as ascertained by the MOVIE metric \cite{movie}
     
     \textbf{4.} Exhaustive experiments on SVHN, Standford Cars, CelebA image dataset and VidTIMID video dataset manifest the benefits of our approach
%=======================================
\section{GAN preliminaries}
A GAN model consists of two deep neural nets, viz., generator, $G$, and discriminator, $D$. The task of the generator is to create an image, $x\in \mathcal{R}^{H\times W \times 3}$ with a noise prior vector, $z\in \mathcal{R}^d$, as input. $z$ is sampled from a known distribution, $p_z(z)$; usually $z\sim \mathcal{U}[-1,1]^d$. The discriminator has to distinguish between real samples(sampled from real distribution, $p_{data}$) and generated samples. The game is played on $V(D,G)$: 
$$ \underset{G}{min} \underset{D}{max} V(D, G) =\mathbb{E}_{x \sim p_{data}(x)}[ log D(x)]$$
\begin{equation}
+\mathbb{E}_{z\sim p_{z}(z)}[1 - D(G(z))].
\label{eq_gan_main_goodfellow}
\end{equation}
%===========================
\section{Method}
We build upon the unsupervised inpainting framework of Yeh \text{et al.}\cite{yeh2017semantic}. Given a masked image, $I_d = M \odot I$, corresponding to an original image, $I$, and a pre-trained GAN model, the idea is to iteratively find the `closest' $z$ vector (starting randomly from $\mathcal{U}[-1,1]^d$) which results in a reconstructed image whose semantics are similar to corrupted image. $z$ is optimized as,
\begin{equation}
\hat{z} = \underset{z}{\mathrm{argmin}}~~ J(M \odot G(z), I_d).
    \label{eq_yeh_loss}
\end{equation}
where $M$ is the binary mask with zeros on masked region else unity, $\odot$ is the pointwise multiplier and  $J(\cdot)$ is the objective function to be minimized. Interesting to note is that the objective function never assumes knowledge of pixel intensities inside the masked region(and thus the term `unsupervised'). Upon convergence, the inpainted image, $\hat{I}$, is given as, $\hat{I} = M \odot I + (1-M) \ \odot G(\hat{z})$. The objective function, $J(\cdot)$ is composed of two components:\\
\textbf{Fidelity Loss:} This loss ensures that the predicted noise prior preserves fidelity between generated image and the original unmasked regions.
\begin{equation}
L_{f} = |M \odot (I - G_{\theta_G}(\hat{z}))|
\end{equation}
\textbf{Perceptual Loss:} This loss ensures that the inpainted output lies near the original/real data manifold and is measured by the log likelihood of real class assigned by the pre-trained discriminator;
\begin{equation}
L_{p} = \log(1 - D_{\theta_d}(G_{\theta_G}(\hat{z})) 
\end{equation}
The overall objective, $J(\cdot) = L_{f} + \lambda L_{p}$, where $\lambda$ controls the relative importance of $L_p$.
\subsection{Better initiation for noise prior search}
One of the fundamental drawbacks of Yeh \textit{et al.} is the iterative optimization requirement of Eq.\ref{eq_yeh_loss}. A random $z \sim \mathcal{U}[-1,1]^{d}$ usually tends to generate images quite disparate from the concerned maksed image and thus the optimization requires multiple updates of $z$. In fact, the authors in 
\cite{yeh2017semantic} suggest around 1000 rounds of iterations per image. Our motivation is to initiate $z$ by respecting some global statistics of the concerned masked image.\\
\textbf{Nearest neighbor search:} After training a GAN, we store (one time offline task) a pool, $P$, of $N$ images by passing $N$ random noise vectors through the pre-trained generator. For a given damaged image, $I_d$, we perform a nearest neighbor search over the pool, $P$, to identify `the closest' matching pair. Specifically, we perform the matching between $I_d$ and a candidate pooled image (generated from $z_i$), $p_i = G(z_i)$, based on a distance metric, $D(I_d, M \odot p_i)$. Please note, even during matching we are not exploiting the masked region of $I_d$. While formulating $D(\cdot)$ we want to make sure that we not only match the overall color statistics of the damaged image but also respect the overall structure. Thus $D(\cdot)$ has got two components:\\
\textbf{Data loss:} This loss penalizes if the pixel intensities of a pooled image, $p_i$, deviate from the damaged image, $I_d$;
\begin{equation}
    L_D = |I_d - (M \odot p_i)|.
    \label{eq_data_loss}
\end{equation}
\textbf{Structure loss:} This loss penalizes if the structure(captured in essence with gradients) of pooled image deviates from damaged image. Structure loss, $L_S$ is defined as:
\begin{equation}
L_S = |\nabla_x I_d - \nabla_x M \odot p_i| + |\nabla_y I_d - \nabla_y M \odot p_i|,
\label{eq_structure_loss}
\end{equation}
where $\nabla_x $ and $\nabla_y$ are horizontal and vertical gradient operators.
The final matching criterion is, $L_{nn} = L_D + \gamma L_S$, where $\gamma$ controls relative importance of $L_S$. Effect of $\gamma$ is discussed in Fig.\ref{fig_data_vs_edge}. The initial noise vector, $z_{init}$ is given by,
\begin{equation}
    z_{init} = \underset{z}{\mathrm{argmin}}~~ L_{nn}(I_d, G(z)); z \in \{z_0, z_1,..., z_N\}
    \label{eq_nn}
\end{equation}
%===========================================
\subsection{Video Inpainting: Exploiting temporal redundancy}
To our best knowledge, this is the first demonstration of unsupervised GAN based inpainting on videos. By video inpainting we refer to concept of error concealment in video, i.e., to recover damaged/masked portion of a frame. A naive application of \cite{yeh2017semantic} would be to apply single image model independently on each frame. This poses two problems, viz, a) such approach does not leverage temporal redundancy among neighboring frames b) independent frame level reconstructions result in temporal inconsistency in a sequence. We propose to address these challenges with two innovations.\\
\textbf{Reuse of predicted $z$ vector:} It is safe to assume that neighboring frames are coherent in appearance and thus the noise priors. Thus, it makes sense to initiate $z_{t+1} = z_t$. This drastically speeds up optimization by almost 80$\times$ compared to vanilla version of Yeh \textit{et al.}\cite{yeh2017semantic}. We refer to this proposed method as \textit{Proposed (Re)} in all experiments.\\
\textbf{Group consistency loss:} Even though we initialize $(z_{t+1})$ with $z_t$, the final solution for time step ($t+k$) can diverge away appreciably from $z_t$. This would mean that there will abrupt changes in scene appearance when the frames are viewed as a sequence. To enforce smooth temporal dynamics we impose a group consistency loss, $(L_G)$, by constraining a group of reconstructed frames to be similar. Disparity between two generated images can be expressed with corresponding disparity between the corresponding $z$ vectors \cite{zhu2016generative}. Specifically, we impose consistency loss over a window of $W$ neighboring frames,
\begin{equation}
    L_G = |z_i - z_k|; ~\forall i \in [1, W],~ \forall k \in [1, W]
    \label{eq_group_consistency}
\end{equation}
Please note that $L_G$ is imposed only over a neighboring window of $W$ frames and not over entire video. Combination of this loss + reuse of $z$ vector is denoted as \textit{Proposed (Re + G)} in experiments. 
%===============================================
\section{Experiment Settings}
\begin{figure}[!t]
 \centering
 \includegraphics[scale = 0.38]{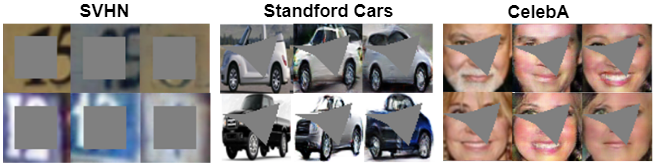}
 \caption{\scriptsize Role of data loss, $L_D$ (Eq.\ref{eq_data_loss}) and structural loss, $L_S$(Eq.\ref{eq_structure_loss}) on retrieving nearest matching initial solution. For each tuple, left column: masked image, middle column: initial solution retrieved with only $L_S$ , right column: initial solution retrieved with only $L_D$. Using $L_D$ mainly tries to maintain the global color statistics while only $L_S$ focuses on matching the structure irrespective of absolute intensity concern. It can be appreciated that only $L_S$ retrieves initial matches by maintaining facial expression(smile), orientation of cars, keystrokes of digits. Thus we apply weighted combination of $L_D$ and $L_S$ as objective (Eq.\ref{eq_nn}) for nearest neighbor retrieval.}   \label{fig_data_vs_edge}
 \end{figure}

\begin{figure}[!t]
 \includegraphics[scale = 0.38]{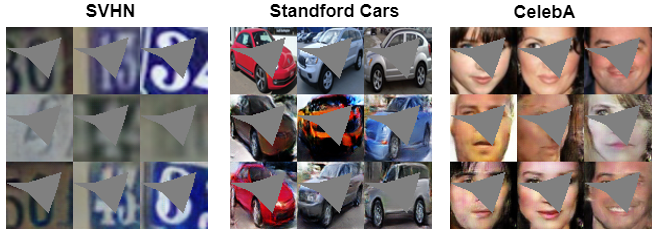}
 \caption{\scriptsize Benefit of proposed noise prior initialization v/s random initialization of Yeh \textit{et al.} \cite{yeh2017semantic}. Top row: Masked image, Middle row: initial solution of \cite{yeh2017semantic}, Bottom row: initial solution by our proposed nearest neighbor based. It is evident that our initial solutions are much more closer to masked images and thus requires lesser iterative updates of Eq.\ref{eq_yeh_loss}. compared to \cite{yeh2017semantic}. }  
 \label{fig_init_benefit}
 \end{figure}

\textbf{Datasets:} For image inpainting we tested on SVHN\cite{netzer2011reading}, Standford Cars\cite{cars} @ 64$\times$64 resolution and CelebA\cite{liu2015deep} @ 64$\times$64 and 128$\times$128 resolution. For video inpainting, we experimented at 128$\times$128 resolution on VidTIMIT\cite{vidtimit} dataset.\\
%=================
\textbf{Network Architectures:} For fair comparison with our baseline of \cite{yeh2017semantic}, we borrowed their architectures for both generator and discriminator and followed their paradigm of training GAN. Parameter, $\lambda$, was set to 0.01 following \cite{yeh2017semantic}.\\
%===========================
\textbf{Balacing data loss and structure loss:}
 Hyperparameter, $\gamma$ sets the relative importance of structure loss, $L_S$, over data loss $L_D$. Setting $\gamma =0$ means initial solution will not explicitly preserve edge information and just retrieve nearest image based on raw intensity. On contrary, a high value of $\gamma$ will enforce only edge preservation without respecting the intensity. See Fig.\ref{fig_data_vs_edge} for illustration of these two extreme cases. In either of these cases, initial solution can be appreciably different than desired solution and will thus require longer iterative refinements. We reserved a validation dataset on which we test $\gamma \in \{0.001, 0.005, 0.01, 0.05, 0.1, 0.5\}$. $\gamma = 0.01-0.03$ gave peak speedup across datasets and thus $\gamma$ is set to 0.01 for all experiments. See Fig.\ref{fig_init_benefit} for some examples of initial solutions with our proposed method v/s random initialization of \cite{yeh2017semantic}.\\
%===================================
\textbf{Pool size ($N$) for nearest neighbor search:}
Nearest neighbor search with an ideal generator ($p_G = p_{data}$) will be able to retrieve the exact $z$ vector corresponding to a masked image, $I_d$, if we allow, $N$ $\rightarrow$  $~ \infty$. However, for practical viability it is not possible to search over every possible $z$ vector. On validation set, we experiment with different $N$ and compare the average MS-SSIM between initial masked solution and masked image. $N$ is set to 300(on all datasets unless otherwise stated), above which, the MS-SSIM does not increase appreciably.  \\
\textbf{Selecting group consistency window, $W$:}
Setting $W$ (in Eq.\ref{eq_group_consistency}) to a large value results in over smoothing of a sequence in temporal dimension and thus leads to degraded MOVIE metric\cite{movie} due to poor perceptual quality. $W=0$, on the other hand will have no effect in incorporating temporal coherence which again hurts MOVIE metric. On held out validation set of VidTIMIT we experimented with $W \in [1, 10]$. $W=5$ was selected which, on average, yielded best MOVIE metric.
So, for a sequence, every alternate 5$^{th}$ frame (pivot frame) is inpainted as a single image. Intermediate frames are initiated with inpainted pivot frame and solutions of a group are additionally constrained by $L_G$.\\
%===============================================
\textbf{Comparing Methods:} Our primary comparing method is the unsupervised baseline of \cite{yeh2017semantic}. However, for the completeness of the paper we also provide comparisons with hybrid frameworks of \cite{context_encoders, iizuka2017globally, gip}. Please note, the later approaches are only successful if trained initially with supervised reconstruction loss. If trained with only unsupervised adversarial loss, these method fail drastically.
%=============================================
 %=======================================
 %============fig_init_benefit starts====
 %======================= figure fig_init_benefit ends====
 %===========================================
 \begin{table}[!h]
 \scriptsize
\caption{\scriptsize Comparing inpainting PSNR (in dB) on different datasets with the unsupervised baseline of \cite{yeh2017semantic}. We also compare with hybrid methods of \cite{context_encoders,iizuka2017globally, gip}.}
\centering 
\begin{tabular}{llllllc}\\\hline
                                & \cite{context_encoders}                       & \cite{iizuka2017globally}                    & \cite{gip}                      & \cite{yeh2017semantic}                      & \begin{tabular}[c]{@{}c@{}}Ours \end{tabular} \\\hline
Cars                            & 14.3                     & 15.3                     & 14.5                     & 13.5                     & 14.1  \\
SVHN & 21.5  & 23.6   & 23.7  & 20.4  & 22.0 \\
CelebA(64) & 23.0 & 24.1  & 24.2 & 22.6 &23.3 \\
\multicolumn{1}{l}{CelebA(128)} & \multicolumn{1}{l}{20.0} & \multicolumn{1}{l}{20.9} & \multicolumn{1}{l}{20.6} & \multicolumn{1}{l}{17.6} & \multicolumn{1}{l}{18.8}                                \\\hline
\label{table_psnr} 
\end{tabular}
\end{table} 
%==========================================
%==========================================
\begin{table}[!t]
\scriptsize
\centering
\caption{\scriptsize Comparison of absolute run times (in seconds) on a NVIDIA K-40 GPU on image (64$\times$64) and video (128$\times$128) inpainting. Time is measured till corresponding loss of a model converges to 95\% of saturation value.  Notice how application of single image model of \cite{yeh2017semantic} naively on higher resolution video booms up the inference time; however, our proposed model (Re + G) appreciably speeds up by almost 80\%. In fact, on image and video we achieve speedups 5$\times$ and $100\times$ in terms of iterations count. This table also considers the time of nearest neighbor search.}
\label{table_time-comparison}
\begin{tabular}{lll}\\\hline
         & Image & Video \\
Yeh \textit{et al.}\cite{yeh2017semantic}      & 9.0     & 33.5  \\
Proposed (Re) & 1.9     & 0.36\\
Proposed (Re + G) & -    & 0.41 \\\hline
\end{tabular}
\end{table}
%=============================================
%=============================================
\begin{table}[!t]
 \scriptsize
\caption{\scriptsize Average temporal consistency ($\eta$) in dB on test sets of different dataset. Higher value means a model is more temporally coherent.}
\centering 
\begin{tabular}{llllllc}\\\hline
                                & \cite{context_encoders}                       & \cite{iizuka2017globally}                    & \cite{gip}                      & \cite{yeh2017semantic}                      & \begin{tabular}[c]{@{}c@{}}Ours\\ (Re)\end{tabular} & \begin{tabular}[c]{@{}c@{}}Ours\\ (Re + G)\end{tabular} \\\hline
Cars                            & 13.8                     & 14.2                     & 14.5                     & 15.6                     & 18.2                                                    & 20.0                                                        \\
SVHN                            & 21.3                     & 21.8                     & 22.1                     & 22.8                     & 24.1                                                    & 24.8                                                        \\
CelebA(64)                      & 23.1                     & 23.2                     & 23.6                     & 24.1                     & 25.6                                                    & 26.3                                                        \\
\multicolumn{1}{l}{CelebA(128)} & \multicolumn{1}{l}{21.8} & \multicolumn{1}{l}{20.9} & \multicolumn{1}{l}{21.6} & \multicolumn{1}{l}{21.9} & \multicolumn{1}{l}{22.4}                                & \multicolumn{1}{c}{23.5} \\\hline         \label{tab_consistency}             
\end{tabular}
\end{table}
%==============================================
\begin{table}[!t]
\scriptsize
\caption{\scriptsize Comparison on MOVIE metric\cite{movie} $\in[0,1]$ on ViDTIMIT video test set. We compare with unsupervised baseline of Yeh \textit{et al.} \cite{yeh2017semantic} and also with hybrid methods of \cite{context_encoders, iizuka2017globally, gip}. A lower MOVIE metric is better in terms of spatio-temporal effectiveness of inpainting.}
\label{tabel_movide_metric}
\begin{tabular}{llllcc}\\\hline\hline
\cite{yeh2017semantic}    & \cite{context_encoders}   & \cite{iizuka2017globally}    & \cite{gip}    & Proposed (Re) & Proposed (Re + G) \\
0.66 & 0.63 & 0.55 & 0.46 & 0.52          & 0.47\\\hline             
\end{tabular}
\end{table}
\section{Results}
\begin{figure}[!t] 
     \centering
     \includegraphics[scale=0.4]{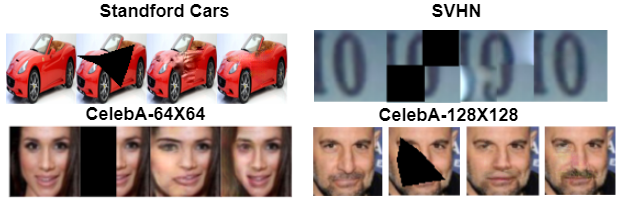}
     \caption{\scriptsize  Visual comparison of inpainting by the unsupervised baseline of \cite{yeh2017semantic}(3$^{rd}$ column) and our proposed method(4$^{th}$ column). We perform equivalently(sometimes better) with 5$\times$ less iterations. 1$^{st}$ col: Original; 2$^{nd}$: masked image.}
     \label{fig_visual_compare_yeh}
 \end{figure}

\begin{figure}[!t]
     \centering
     \includegraphics[scale=0.35]{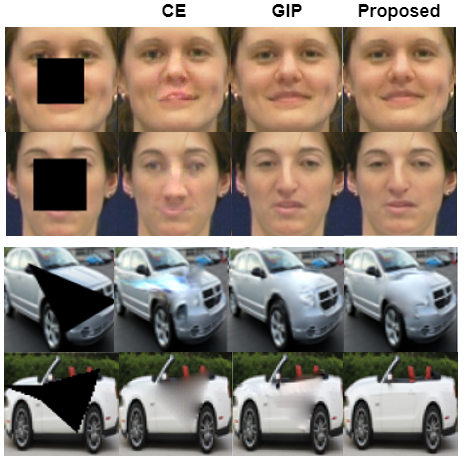}
     \caption{\scriptsize Comparing with contemporary hybrid methods of CE\cite{context_encoders} and GIP\cite{gip}.}
     \label{fig_visual_comp_others}
 \end{figure}

\begin{figure}[!h]
 \centering
 \includegraphics[scale = 0.6]{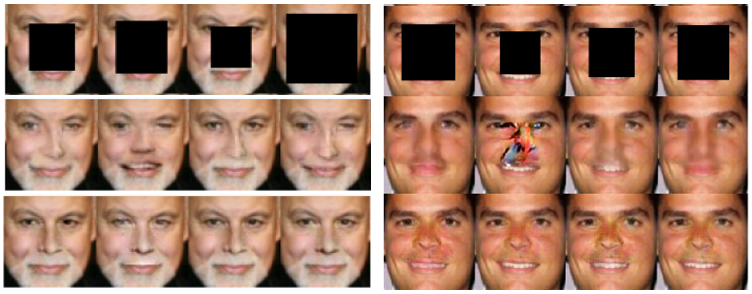}
 \caption{\scriptsize Visualizing benefit of \textbf{proposed model} (bottom row) over Yeh \textit{et al.\cite{yeh2017semantic}} (middle row) on pseudo sequences. A pseudo sequence is created from a single image of a person but masked with different masks (thereby mimicking a temporal aspect). An ideal sequence inpainting model should result in identical outputs for a given subject. Note that our sequence reconstructions lead to more temporally consistent (notice the lips, eyes) solutions.}  
 \label{fig_consistency}
 \end{figure}

\textbf{Speedup in optimization:} With respect to our unsupervised baseline\cite{yeh2017semantic}, on average, we achieved about 5$\times$ speedup for single image inpainting. On videos the speedup is almost 80$\times$. See Table \ref{table_time-comparison} for speed comparisons.\\
\textbf{Image Inpainting:} In Fig. \ref{fig_visual_compare_yeh} we show some exemplary inpainting comparison with \cite{yeh2017semantic}. Even with appreciable speedup we usually achieve better (or similar visual performance) to \cite{yeh2017semantic}. We also show some visual comparison with recent hybrid benchmarks in Fig.\ref{fig_visual_comp_others}. Recently \cite{srgan, yeh2017semantic, gip} researchers have shown that PSNR metric is not fully justifiable to assess tasks such as inpainting and super resolution. However, for reference we also report PSNR in Table\ref{table_psnr}.\\ 
%==========================
\textbf{Pseudo sequences and temporal consistency}
Before analyzing effects of our proposed losses on real videos, we study the effects on a simpler case of pseudo sequences. A pseudo sequence of length, $S$ is basically a single image replicated $S$ times but masked with different masks. An ideal model would inpaint all the frames identically. We can define temporal consistency, $\eta$ as, $\eta = \frac{1}{{S\choose 2}} \sum_{k=1}^S  PSNR(\Hat{I}_k^i, \Hat{I}_j^i)~ \forall$ combinations of $(j, k) \in [1, 2, ... S]$.
%\begin{equation}
%    \eta = \frac{1}{S}\sum_{k=1}^S  PSNR(I_d^k, \Hat{I}^i)
%    \label{eq_temporal_consistency}
%\end{equation}
In Table \ref{tab_consistency} we report temporal consistency over different datasets. It can be seen that proposed initialization technique of $z_{t+1}=z_t$ improves consistency with further improvement brought by group consistency loss. Even the current hybrid benchmarks \cite{gip, context_encoders, iizuka2017globally} manifest greater inconsistency because these methods do not leverage any temporal information. Exemplary visualizations are shown in Fig. \ref{fig_consistency}. Note, $\eta$ gives an indication of temporal consistency only (which can be studied only on such pseudo sequences). It does not give essence of spatial correctness. For example, a model might reconstruct all blank images which will give high $\eta$, but will yield worse MOVIE metric.\\
%=======================================
\textbf{Analyzing real video inpainting performance} 
PSNR and MS-SSIM are not well suited for judging a reconstructed video since these metrics are agnostic to temporal dimension. We advocate using the MOVIE metric\cite{movie} which considers spatial, temporal and spatio-temporal aspects by comparing original and reconstructed sequence. In Table \ref{tabel_movide_metric} we report average MOVIE metric on VidTIMIT test set. Our proposed modifications to \cite{yeh2017semantic} significantly improves reconstructed video quality.
%===========================================
%=====================================
 %===================================
 %==========================================
 %=======================================
 \vspace{-9mm}
 %===========================
\section{Conclusion}
In this paper, we first discussed the problem of impractical long inference time of the recent  completely unsupervised inpainting framework of \cite{yeh2017semantic}. We then proposed to speedup the iterative optimization of \cite{yeh2017semantic} by better initialization technique on images and also leveraging temporal redundancy in videos. In the process, we achieved a significant speedup. Several comparisons were also done with current hybrid benchmarks and we achieved comparable performance. Future work might consider replacing the iterative optimization by learning to project a damaged image into noise prior space.
%================== end of main ==========
\bibliographystyle{IEEEbib}
\bibliography{main}
\end{document}